\documentclass[lettersize,journal]{IEEEtran}
\usepackage{amsmath,amsfonts}
\usepackage{array}
\usepackage[caption=false,font=normalsize,labelfont=sf,textfont=sf]{subfig}
\usepackage{textcomp}
\usepackage{stfloats}
\usepackage{url}
\usepackage{graphicx}
\usepackage{cite}
\usepackage{caption}

\usepackage{xcolor} 
\usepackage{quiver}
\usepackage{amsthm}
\usepackage{listings}
\usepackage{mathtools}
\usepackage[linesnumbered,ruled,vlined]{algorithm2e}
\usepackage[T1]{fontenc}
\usepackage{tgcursor}

\usepackage[frozencache,cachedir=minted-cache]{minted}
\setminted{xleftmargin=1em, escapeinside=||}
\newcommand*{\jul}[1]{\mintinline{text}{#1}}

\newcommand{\cat}[1]{\mathsf{#1}}
\newcommand{\Set}{\mathsf{Set}}
\newcommand{\catSet}[1]{\cat{#1}\text{-}\Set}

\newtheorem{definition}{Definition}

\begin{document}

\title{Automating Transfer of Robot Task Plans using Functorial Data Migrations}



\author{Angeline Aguinaldo
\IEEEmembership{Member, IEEE}, Evan Patterson, William Regli
\IEEEmembership{Fellow, IEEE}
\thanks{A. Aguinaldo and W. Regli are with the department of Computer Science, University of Maryland, College Park, MD, USA (e-mail: aaguinal@umd.edu, regli@umd.edu)}
\thanks{E. Patterson is a Research Scientist at the Topos Institute, Berkeley, CA, USA (e-mail: evan@topos.institute)}
}




\thispagestyle{empty} 
\textcopyright\ 2025 IEEE. Personal use of this material is permitted. Permission from IEEE must be obtained for all other uses, in any current or future media, including reprinting/republishing this material for advertising or promotional purposes, creating new collective works, for resale or redistribution to servers or lists, or reuse of any copyrighted component of this work in other works.
\newpage

\maketitle

\begin{abstract}
  This paper introduces a novel approach to ontology-based robot plan transfer by leveraging functorial data migrations, a structured mapping method derived from category theory. Functors provide structured maps between planning domain ontologies which enables the transfer of task plans without the need for replanning. Unlike methods tailored to specific plans, our framework applies universally within the source domain once a structured map is defined. We demonstrate this approach by transferring a task plan from the canonical Blocksworld domain to one compatible with the AI2-THOR Kitchen environment. Additionally, we discuss practical limitations, propose benchmarks for evaluating symbolic plan transfer methods, and outline future directions for scaling this approach.

  \emph{Note to Practitioners}---This research was motivated by the challenge of reusing successful robot task plans in different domains without the need to redesign or recalibrate the plans from scratch. In industries where automation is prevalent, such as manufacturing or logistics, the ability to efficiently transfer a task plan from one setting to another can significantly enhance productivity and reduce operational costs. Our solution utilizes functorial data migrations, a concept from category theory, to enable the structured transfer of task plans between distinct domains. This approach ensures that transferred plans remain valid and applicable. For example, a plan developed to transport items in a warehouse can be adapted to transport medical supplies in a hospital, which leverages the underlying structure of the tasks without altering their essence. While our approach generalizes task plan transfer across domains, it relies on precise domain ontology maps, which can be complex. This method may not capture all nuanced transfers between highly-specialized domains, particularly where domain-specific knowledge is comprehensive. Practitioners should be aware that some manual tuning might still be necessary to fully optimize the plans for specific operational contexts.
\end{abstract}



\section{Introduction}
\IEEEPARstart{R}{obots} are becoming increasingly capable of computing short-horizon motion plans either by way of motion planners, teach pendants, or other programming methods. These short plans, however, are not suited to complete the lofty goals in the variety of environments we hoped they would. This has led to successful, albeit tailored, methods for composing atomic gestures into longer plans to execute more complex tasks \cite{Mansouri2021,Lagriffoul2018,Guo2023,Garrett2021}. Tailoring these plans to perform well in an environment is a delicate endeavor that requires careful engineering of the description of the task, among other things. After such an investment, there is a natural desire to seek out methods for translating plans so that they can be reused in new environments. 


\begin{figure}
  \centering
  \includegraphics[width=\columnwidth]{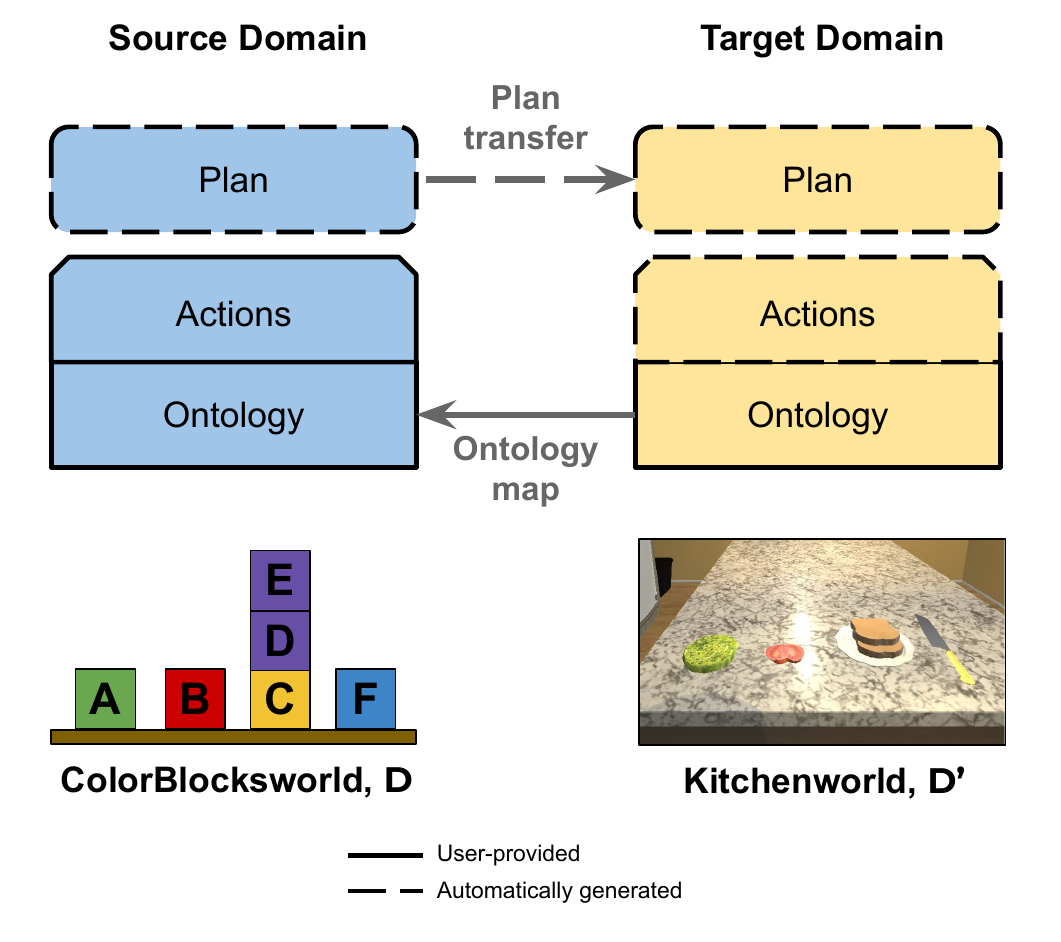}
  \caption{A conceptual illustration of transferring a plan from ColorBlocksworld to AI2-THOR Kitchenworld, using an ontology map to formalize the translation between planning domains and automatically transfer a valid plan to the target domain.}
  \label{fig:concept}
\end{figure}

There are a few obstacles inherent to task representations that must be addressed before realizing generalized plan transfer. Task plans use symbolic representations of the world, as opposed to continuous ones in order to reason about complex goals. The ontology that serves as the basis of these symbolic representations requires significant knowledge engineering effort. For every new setting, application experts must re-design the language with which to talk about the world and solve for a plan again, even if the tasks appear analogous or similar. The ontology also influences how discrete action models are defined in this setting. This means that even a slight difference in the domain ontology could have a large impact on the action models, requiring one to replan. By defaulting to replanning, a user may incur long planning times, produce plans that are not proper analogies and, ultimately, ignore generalizations suggested by their intuition.

Related work addressing this challenge largely focuses on transfer learning of skills in robotics \cite{Elimelech2023SPAR1,Elimelech2023SPAR2,Men2023,Morere2019,Heuss2023,Jin2024,Zhao2022,Pitkevich2024,Hao2022,Jin2025}. However, these methods are often narrowly tailored to specific use cases and fail to generalize effectively across diverse domains. Additionally, they lack the formal guarantees required for consistent and valid plan transfer. In contrast, our method leverages functorial data migration, a formal framework rooted in category theory, to enable well-defined mappings between planning domain ontologies. By structuring domain ontologies as categories, category-theoretic principles provide clear guidelines for constructing new ontologies and actions to ensure the transfer process is mathematically sound.  Consequently, this approach enables the migration of actions from one domain to another, automatically generating analogous plans in the target domain without the need for replanning.

In summary, the main contributions of this paper are:

\begin{itemize}
  \item A novel application of functorial data migration from category theory to robotic task planning,
  \item A novel algorithm for automating plan transfer between distinct planning domains using ontology mappings, and
  \item A case study demonstrating the success of the approach for adapting actions in a generic domain, namely Blocksworld, to a richer application, namely one compatible with the AI2-THOR \cite{Kolve2017} Kitchen environment
\end{itemize}

In the remainder of this paper, we will introduce the motivating example, provide background on task planning and category theory, describe the concepts behind the domain ontology maps and plan transfer, and apply this approach to a transfer scenario involving a variation of the canonical planning domain, Blocksworld, and a kitchen simulation environment.

\section{Example: Blocksworld to Kitchenworld}
Transferring robot task plans from a variation of the Blocksworld domain to a richer planning domain holds significant practical relevance for real-world applications. In Blocksworld, tasks are oriented around arranging simple blocks in specific configurations. This task is analogous to many industrial and home automation tasks. For example, one clear analogy could be found when considering how to arrange ingredients to construct a sandwich. This task can be better expressed in a domain such as AI2-THOR Kitchenworld. So, while Blocksworld has the benefit of generality, Kitchenworld has the benefit of rich expressivity necessary for real applications. During planning, however, rich states can hinder computation, therefore it is sometimes favorable to plan in an abstracted domain and translate the plan to a richer one. In Section \ref{sec:case}, we will elaborate on a case study of this example, demonstrating how a task plan can be transferred to achieve this outcome using our approach.

\section{Background}

\subsection{Planning Domain}

Task planning in robotics is concerned with finding a sequence of high-level actions that transition an initial state to a goal state. A task planning scenario can be broken down to the \emph{planning domain} and the \emph{planning problem} \cite{Ghallab2004}. The planning domain refers to the language used to talk about the world and what can be done in it. We can distinguish parts of the planning domain as being the \emph{domain ontology} and the \emph{domain actions}. The planning problem specifies the initial and goal states and requires a sequence of domain actions to connect them.

The domain ontology specifies the \emph{types} and \emph{predicates}. Types are used to classify entities in the planning domain. Predicates are used to define relations between these types. An example of such a specification can be found in the preamble of a PDDL (Planning Domain Definition Language) domain file \cite{pddl4j_blocksworld}. For example, the Blocksworld domain introduces the predicates that can be used to describe features of the state, such as a block being on another block \jul{(on ?x - block ?y - block)} or a block being on the table \jul{(ontable ?x - block)}. The annotations (\jul{- block}) indicate that \jul{?x} and \jul{?y} have type \jul{block}. 

The domain actions are formal descriptions of the operations that can be performed within a planning domain. STRIPS \cite{Fikes1971} is a widely-accepted model for specifying symbolic actions, which describes actions in terms of \emph{preconditions} and \emph{effects}. Preconditions are conditions that must be true in the world before the action can be executed. Effects are conditions that are true in the world after the action is executed. Both the preconditions and effects are expressed in the language of the domain ontology.

Together, the domain ontology and actions form the planning domain. Because of this, it is easy to see that the domain is greatly influenced by its ontology design. Minor changes to the domain ontology can yield large changes to the domain actions in order to suit the semantics of the domain.

\subsection{Category Theory Fundamentals}
Categories, functors, and natural transformations are fundamental structures presented in category theory---from which many other structures are derived. These mathematical structures serve as the basis for our task planning formalism and plan transfer framework. Here, we will provide an expedient introduction to the fundamentals, however, a detailed treatment can be found in \cite{Leinster2016}. 

A \emph{category}, $\cat{C}$, is a mathematical object that is defined in terms of \emph{objects} and \emph{morphisms}. Morphisms are arrows between two objects that, in many cases, represent functions and relations. For example, $f: X \rightarrow Y$ denotes a morphism from objects $X$ to $Y$ in $\cat{C}$. Morphisms are composable if the target of one morphism matches the source of another. The composition operation is often denoted as $g \circ f$, which can be read as "$f$ then $g$". Lastly, for every object in a category, there exists an identity morphism, $1_X$, that sends an object to itself.

A \emph{functor} is a map from one category to another. For example, $F: \cat{C} \rightarrow \cat{D}$ denotes a functor from category $\cat{C}$ to category $\cat{D}$. The functor maps objects from the source category, in this case $\cat{C}$, to objects in the target category, in this case $\cat{D}$. The functor also does the same for the morphisms, preserving composition and identities.

A \emph{natural transformation} is a way to translate one functor to another while preserving the structure within the source category, say $\cat{C}$. Specifically, this means that for functors $F, G: \cat{C} \rightarrow \cat{D}$, a natural transformation, $\alpha: F \rightarrow G$, consists of a family of component maps such that, for any morphism $f: X \rightarrow Y \in \cat{C}$, a commuting square in $\cat{D}$ is formed with the components $F(f)$, $G(f)$, $\alpha_X: F(X) \rightarrow G(X)$ and $\alpha_Y: F(Y) \rightarrow G(Y)$. 

\subsection{$\catSet{C}$ and Double-Pushout Rewriting}

In previous work \cite{Aguinaldo2023}, we demonstrate that functors and categories of functors can be used to model states and actions in planning. The functor we will focus on for this work is $\cat{C} \rightarrow \cat{Set}$, which has also been shown to be an elegant formal semantics for relational databases \cite{Spivak2012}.

\begin{definition}
A $\cat{C}\text{-}\mathrm{set}$ is a functor from the $\cat{C}$ to the category $\cat{Set}$. Here $\cat{C}$ is a finitely presented category which is often referred to as the \textbf{schema category} and $\cat{Set}$ is a category whose objects are sets and whose morphisms are set functions.
\end{definition}

Finitely presented refers the fact that objects and morphisms are generated from finite sets of objects and morphisms. The objects and morphisms in a schema category, for our purposes, define the types and predicates in a domain ontology. The functor, $\cat{C}\text{-}\mathrm{set}$, identifies a set for each type and a set function for each predicate. By convention, these sets are finite. 

$\cat{C} \rightarrow \cat{Set}$ functors themselves can exist within a category, which we call the category $\catSet{C}$.

\begin{definition}
The category $\catSet{C}$ is a category whose objects are $\cat{C} \rightarrow \cat{Set}$ functors and whose morphisms are natural transformations. 
\end{definition}

A number of shapes can appear when looking at objects and morphisms in $\catSet{C}$. One, in particular, is a \emph{span} which is a diagram that takes the shape ($\bullet \leftarrow \bullet \rightarrow \bullet$). Certain spans in $\catSet{C}$ can be referred to as \emph{rewrite rules} which are graph rewrite rules generalized for categories \cite{Brown2021}. In this planning formalism, rewrite rules can be interpreted as domain actions in a planning domain \cite{Aguinaldo2023}. More formally:

\begin{definition}
An \textbf{action} is a span in $\catSet{C}$: 

\begin{center}
$\mathrm{Pre} \xhookleftarrow{l} \mathrm{Keep} \xrightarrow{r} \mathrm{Eff}$
\end{center}

\noindent $\mathrm{Pre}$ refers to the precondition of the action and $\mathrm{Eff}$ is the effect of the action. $\mathrm{Keep}$ refers to conditions that remain the same between $\mathrm{Pre}$ and $\mathrm{Eff}$.
\end{definition}

The hooked arrow, $\hookleftarrow$, means that the morphisms are \emph{monic}, also known as \textit{monomorphisms}, which correspond to injective functions $\cat{Set}$. An action can be applied to a state, namely an object $X \in \catSet{C}$, to produce a new state. To ground an action to a state, a \emph{match morphism} must be identified which is a natural transformation $m: \mathrm{Pre} \hookrightarrow X$. The action span and its match can be used to construct a \textit{double-pushout square}.

\[\begin{tikzcd}
	\mathrm{Pre} & \mathrm{Keep} & \mathrm{Eff} \\
	X & Z & Y
	\arrow["l"', hook', from=1-2, to=1-1]
	\arrow["r", from=1-2, to=1-3]
	\arrow[from=1-2, to=2-2]
	\arrow["m"', hook, from=1-1, to=2-1]
	\arrow[from=2-2, to=2-1]
	\arrow[from=2-2, to=2-3]
	\arrow[from=1-3, to=2-3]
	\arrow["\lrcorner"{anchor=center, pos=0.125, rotate=90}, draw=none, from=2-1, to=1-2]
	\arrow["\lrcorner"{anchor=center, pos=0.125, rotate=180}, draw=none, from=2-3, to=1-2]
\end{tikzcd}\]

The result after applying this rule is a state, $Y$, that contains all parts of $\mathrm{Eff}$ and deletes the parts of $\mathrm{Pre}$ that are not identified in $\mathrm{Keep}$. The procedure for producing a new state after applying an action is called \emph{double-pushout (DPO) rewriting}. A thorough explanation of DPO rewriting can be found in Brown et. al. \cite{Brown2021}.

\subsection{Functorial Data Migration}
\label{sec:migration}
Functorial data migration, as conceptualized by Spivak \cite{Spivak2012}, is a mathematical approach to handling and migrating data from a $\catSet{C}$ category to one with a different schema category, e.g. $\cat{C}^{\prime}$. This method uses functors to systematically map data from one schema to another such that the relationships and constraints within the data are preserved. Functorial data migration has been applied across various domains, including database integration \cite{Schultz2017}, model-management in systems engineering \cite{Breiner2019a}, and knowledge representation \cite{Hester2016}. 

There are different types of data migrations \cite{Spivak2012}. One type of data migration is a \textit{delta migration functor}. Formally:

\begin{definition}
  Given a functor $F: \cat{C}^{\prime} \rightarrow \cat{C}$, called a \textbf{translation functor}, a \textbf{delta migration functor}, $\Delta_F: \catSet{C} \rightarrow \catSet{C^{\prime}}$, is constructed as follows. 
  
  For a given object $X$ in $\catSet{C}$,
  
  \;\;\;\; $\Delta_F(X) = X \circ F$
  
  For a given morphism $\alpha: X \rightarrow Y$ in $\catSet{C}$, $\Delta_F$ acts on $\alpha$ to produce the following morphism in $\catSet{C^{\prime}}$---
  
  \;\;\;\; $\Delta_F(\alpha)_{c^{\prime}}: X(F(c^{\prime})) \rightarrow Y(F(c^{\prime}))$ for $c^{\prime} \in \cat{C^{\prime}}$
\end{definition}

A delta migration functor depends on a straightforward, one-to-one correspondence between domain ontologies. 

Another type of data migration is a \textit{conjunctive query migration functor}, which depends on a translation functor that maps types in the target domain to diagrams\footnote{Diagrams are a formal term used in category theory to mean a functor, $\mathrm{Diag}: \cat{J} \rightarrow \cat{C}$, from an indexing shape, $J$, to a category, $\cat{C}$.} of types, or conjunctive queries, in the source domain. 

\begin{figure}[h!]
  \centering
  \includegraphics[width=\columnwidth]{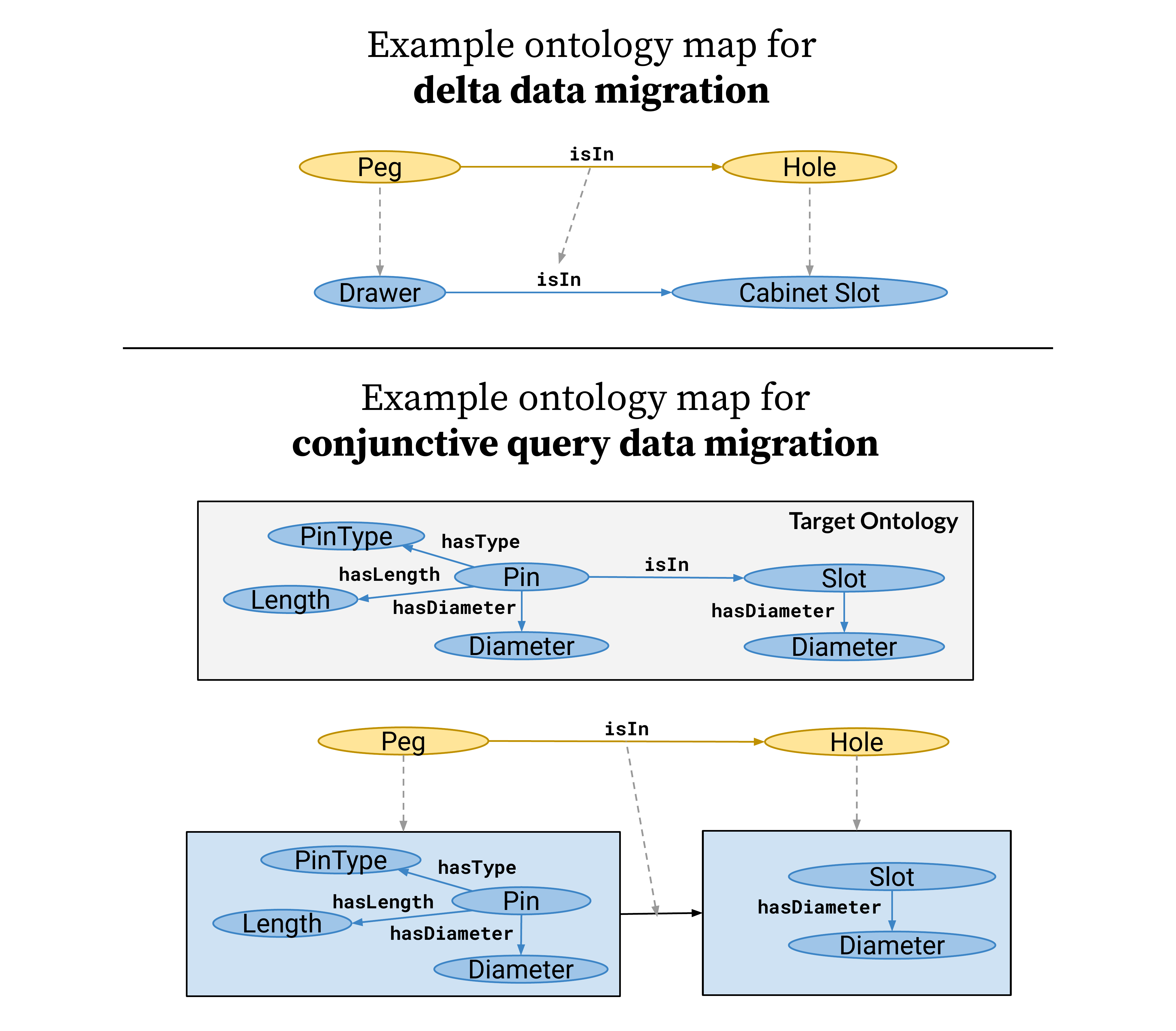}
  \caption{Example of translation functors for a delta data migration (top) and a conjunctive query data migration (bottom).}
  \label{fig:translationfunctorex}
\end{figure}

The distinction between a delta migration functor and a conjunctive query migration functor offers practical advantages, as it allows for flexibility in expressing nuanced correspondences between concepts in different domains. Simple examples of translation functors for a delta migration and conjunctive query migration can be seen in Figure \ref{fig:translationfunctorex}.

\section{Approach}

Plan transfer involves adapting an existing plan to suit a different planning context from the one for which it was originally designed. A key challenge arises when the source and target contexts have differing domain ontologies, which can complicate the direct reuse of plans. In this paper, we address this challenge by focusing on methods for transferring plans across domains with distinct ontological structures. To do so, we begin by introducing a mathematical sketch of the problem, which provides a formal framework for understanding plan transfer, and then elaborate on its individual components.  

\subsection{Problem Formulation}

Recall that a task planning scenario can be broken down into the planning domain, $\textbf{D}$, the planning problem, $\textbf{P}$. The planning domain consists of a domain ontology, $\cat{D}$, and set of domain actions, $A$. The domain actions are operations specified in terms of preconditions and effects. These preconditions and effects start off as being defined with \emph{ungrounded} parameters which means they are bound to types and not to any specific entity in the environment. A \emph{grounded} action is an action whose types are bound to specific entities in the environment. The planning problem can be decomposed into the initial state, $I$, and goal state, $G$. We will denote the planning domain and planning problem as $\textbf{D} = (\cat{D}, A)$ and $\textbf{P} = (I, G)$, respectively. 

A \emph{plan}, $\sigma_{\textbf{D}}$, refers to the sequence of grounded actions within domain $\textbf{D}$. A \emph{valid plan} is a plan that is a solution to $\textbf{P}$, which means there exists an inclusion map from $G$ into the final state of the plan. 

Transferring a plan from one planning domain to another requires defining a map, $F: \cat{D}^{\prime} \rightarrow \cat{D}$, from the target ontology $\cat{D}^{\prime}$ to the source ontology $\cat{D}$. We use $F$ to induce a plan transfer function, $\Delta_{F}$, that maps the original plan, $\sigma_{\textbf{D}}$, to a corresponding plan in the target domain, $\sigma^{\prime}_{\textbf{D}^{\prime}}$. 

\begin{table}[tbp]
  \centering
  \caption{Key between Task Planning and Mathematical Terms}
  \renewcommand{\arraystretch}{1.5} 
  \begin{tabular}{|>{\centering\arraybackslash}m{0.3\columnwidth}|>{\centering\arraybackslash}m{0.6\columnwidth}|}
  \hline
  \textbf{Task Planning} & \textbf{Mathematical} \\
  \hline
  Ontology & Schema category, $\cat{D}$ \\
  \hline
  Domain & $\catSet{D}$ category \\
  \hline
  State & Object $X$ in $\catSet{D}$; functor $X: \cat{D} \rightarrow \cat{Set}$ \\
  \hline
  Ungrounded Actions & Rewrite rules ($\mathrm{Pre} \hookleftarrow \mathrm{Keep} \rightarrow \mathrm{Eff}$) in $\catSet{D}$ \\
  \hline
  Grounded Actions & Rewrite rules in $\catSet{D}$ with match morphism, $m$, from $\mathrm{Pre}$ to another object in $\catSet{D}$ \\
  \hline
  Problem & Pair of objects $(I,G)$ in $\catSet{D}$ \\
  \hline
  Plan & Sequence of rewrite rules with their respective match morphisms \\
  \hline
  Valid Plan & A plan for which an inclusion (monic) morphism $g: G \hookrightarrow Y_n \in \catSet{D}$ exists, where $G$ is the goal and $Y_n$ is the end state\\
  \hline
  Ontology Map & Functor $F: \cat{D^{\prime}} \rightarrow \cat{D}$ \\
  \hline
  Plan Transfer & Data migration functor $\Delta_F$  \\
  \hline
  \end{tabular}
  \label{tab:key}
\end{table}

A \emph{valid plan transfer} is one that maps the original plan, $\sigma_{\textbf{D}}$, to a new plan, $\sigma^{\prime}_{\textbf{D}^{\prime}}$, in the target domain, such that $\sigma^{\prime}_{\textbf{D}^{\prime}}$ is a valid plan for $\textbf{P}^{\prime} = (\Delta_F(I), \Delta_F(G))$ in $\textbf{D}^{\prime}$. This assumes that the initial and goal states in the source domain can be translated to semantically appropriate counterparts in the target domain using $\Delta_F$. 

Considering all the aforementioned, we can summarize plan transfer as having to execute the following high-level steps--

\begin{enumerate}
  \item Define the source domain ontology, $\cat{D}$, for \textbf{D}. 
  \item Instantiate the initial and goal states, $\textbf{P} = (I, G)$, and solve for a plan, $\sigma_{\textbf{D}}$.
  \item Define the target domain ontology, $\cat{D}^{\prime}$, in $\textbf{D}^{\prime}$.
  \item Define the map between the target and source domain ontologies, $F: \cat{D}^{\prime} \rightarrow \cat{D}$.
  \item Transfer the grounded plan, namely $\Delta_F(\sigma_{\textbf{D}}) = \sigma^{\prime}_{\textbf{D}^{\prime}}$.
\end{enumerate}

Table \ref{tab:key} summarizes our interpretation of fundamental concepts in task planning and plan transfer using mathematical terms in category theory.

\subsection{Domain ontologies, $\cat{D}$, and States, $X: \cat{D} \rightarrow \cat{Set}$}
In this approach, the domain ontologies are specified as schema categories, $\cat{D}$ and $\cat{D}^{\prime}$. For each category, the objects represent types, e.g. $T_1, T_2$, and the morphisms represent functional binary predicates, $f: T_1 \rightarrow T_2$, in the domain. 

To create states over this domain, we can ground the types and predicates in $\cat{Set}$ using functor, $X: \cat{D} \rightarrow \cat{Set}$. When grounded in $\cat{Set}$, types can refer to symbols or data primitives, such as strings, integers, and floats. We call sets of data primitives, \emph{attribute types}, and predicates that connect types to attribute types, \emph{attributes}.\footnote{A mathematically rigorous description between types and attribute types can be \cite{Schultz2017,Patterson2021}.}

Recall that a functor preserves composition in the source category. This implies that for every predicate, $f: T_1 \rightarrow T_2$, in $\cat{D}$, there must be a function between the corresponding sets, $X(f): X(T_1) \rightarrow X(T_2)$ in $\cat{Set}$. Additionally, this means that the pair of composable morphisms $f$ and $g: T_2 \rightarrow T_3$ act on the set under $T_1$ the same way the composite morphism, $X(g \circ f): X(T_1) \rightarrow X(T_3)$, acts on the set under $T_1$.

\subsection{Ontology map, $F: \cat{D}^{\prime} \rightarrow \cat{D}$}
The ontology map, $F:  \cat{D}^{\prime} \rightarrow \cat{D}$, is a translation functor (Section \ref{sec:migration}) defined from the target domain, $\cat{D}^{\prime}$, to the source domain, $\cat{D}$. To ensure the functor is well-defined and preserves the structure of the domains, the following rules must be followed when defining a translation functor:
\begin{enumerate}
\item \textbf{Map each type}.  Each type in the target domain must map to a type (or a diagram of types) in the source domain.
\item \textbf{Map each predicate}. Each predicate in the target domain must map to a predicate (or a map between diagrams) in the source domain such that both the corresponding objects and composition are preserved.
\item \textbf{Map each attribute}. Each attribute in the target domain must map to a function that determines its attribute value.  
\end{enumerate}
The design of the ontology map will dictate whether a delta data migration or conjunctive query migration will be applied.



\subsection{Plan transfer functor, $\Delta_F: \catSet{D} \rightarrow \catSet{D'}$}

Provided an ontology map, $F: \cat{D}^{\prime} \rightarrow \cat{D}$, we can define a plan transfer functor, $\Delta_F: \catSet{D} \rightarrow \catSet{D'}$, which corresponds a sequence of actions in $\catSet{D}$ (source plan) to a sequence of actions in $\catSet{D'}$ (transferred plan). 

The objects and morphisms of the category $\catSet{D}$ are used to construct the states and actions within the source planning domain, $\mathbf{D}$. The morphisms of $\catSet{D}$ are used to construct ungrounded actions in $\catSet{D}$ and actions are grounded using match morphisms. For example, the first action in a plan will construct a match morphism from the precondition, $\mathrm{Pre}$, to the initial state, $I$.  

\begin{figure}[t!]
  \centering
  \includegraphics[width=\columnwidth]{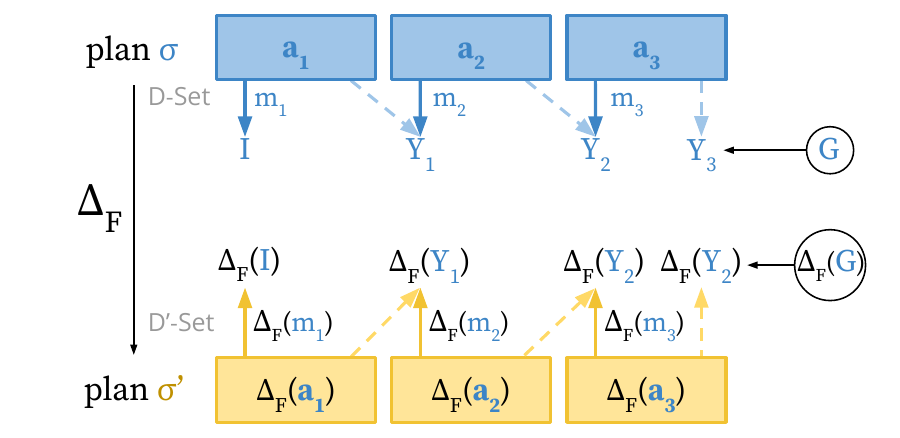}
  \caption{Schematic of plan transfer, $\Delta_F: \catSet{D} \rightarrow \catSet{D^{\prime}}$, for the ontology map, $F: \cat{D^{\prime}} \rightarrow \cat{D}$. The top (blue) sequence represents a plan in $\catSet{D}$. The bottom (yellow) sequence represents a plan in $\catSet{D^{\prime}}$. Rectangles in the schematic correspond to actions within each plan. Arrows emerging from the left side of the rectangles are match morphisms. Dotted arrows emerging from the right side of the rectangles point to the resultant state after each action is applied.}
  \label{fig:transfer}
\end{figure}

When translating plans, each grounded action is transferred using either a delta data migration or a conjunctive query migration functor as defined in Section \ref{sec:migration}. The plan transfer functor will act on all grounded actions in the plan to produce analogous actions in the target domain. The procedure for plan transfer can be viewed algorithmically in Algorithm \ref{alg:transfer}. A schematic of how $\Delta_F$ is applied to a plan can be seen in Figure \ref{fig:transfer}.

\begin{algorithm}[h]
  \SetAlgoLined
  \KwIn{\\
          \Indp \Indp
          Ontology map, $F: \cat{D^{\prime}} \rightarrow \cat{D}$ \\ 
          Source plan, $\sigma$ \\
    }
  \KwOut{Transferred plan $\sigma^{\prime}$}
  
  \SetKwFunction{FMain}{TransferPlan}
  \SetKwProg{Fn}{Function}{:}{}
  \Fn{\FMain{$F$, $\sigma$}}{
      $\sigma^{\prime}$ = [ ]\;
      \ForEach{(\emph{action} $a$, \emph{match} $m$) in $\sigma$}{
          $\Delta_F \leftarrow$ construct plan transfer functor using $F$\;
          $a^{\prime} \leftarrow$ $\Delta_F(a)$\;
          $m^{\prime} \leftarrow$ $\Delta_F(m)$\;
          put ($a^{\prime}, m^{\prime}$) in $\sigma^{\prime}$\;
      }
      \KwRet $\sigma^{\prime}$\;
  }
  \caption{Plan Transfer Procedure}
  \label{alg:transfer}
\end{algorithm} 

\subsection{Validating Plan Transfer}
\label{sec:validate}
To validate the success of plan transfer, it must be true that there is an monomorphism from the corresponding goal state in the target domain, $\Delta_F(G)$, to the end state of the transferred plan. It is well-known that delta data migrations preserve pushout squares \cite{Spivak2012}. As a result, $\Delta_F$ is guaranteed to always produce a valid plan transfer without requiring an additional validation step. A formal proof of this fact is provided in Appendix \ref{sec:proof}. 

In contrast, this fact does not hold for conjunctive query data migrations. Therefore, it is necessary to validate that the transferred plan satisfies the goal in the target domain. Specifically, we must confirm the presence of the monomorphism, $g: \Delta_F(G) \rightarrow Y_{n+1}$. To achieve this, a backtracking search algorithm \cite{Brown2021} can be used to identify such a morphism.


\section{Case Study: ColorBlocksworld to Kitchenworld}
\label{sec:case}
To demonstrate this approach, we consider two domains: a variant of the Blocksworld domain \cite{Russell2021} from classical AI planning, called \emph{ColorBlocksworld}, and an object rearrangement domain, \emph{Kitchenworld}, based on the AI2-THOR simulation environment \cite{Batra2020}. In ColorBlocksworld, the robot is tasked to arrange blocks in a goal configuration while keeping track of what block is in the gripper. Each block has a unique identifier and a potentially non-unique color. In Kitchenworld, the robot is tasked to arrange ingredients and utensils in a goal configuration. These items may serve as receptacles, allowing other items to be stacked on them, and can have attributes such as temperature, mass, and material. 

\begin{figure*}[h!]
  \centering
  \includegraphics[width=\columnwidth * 2]{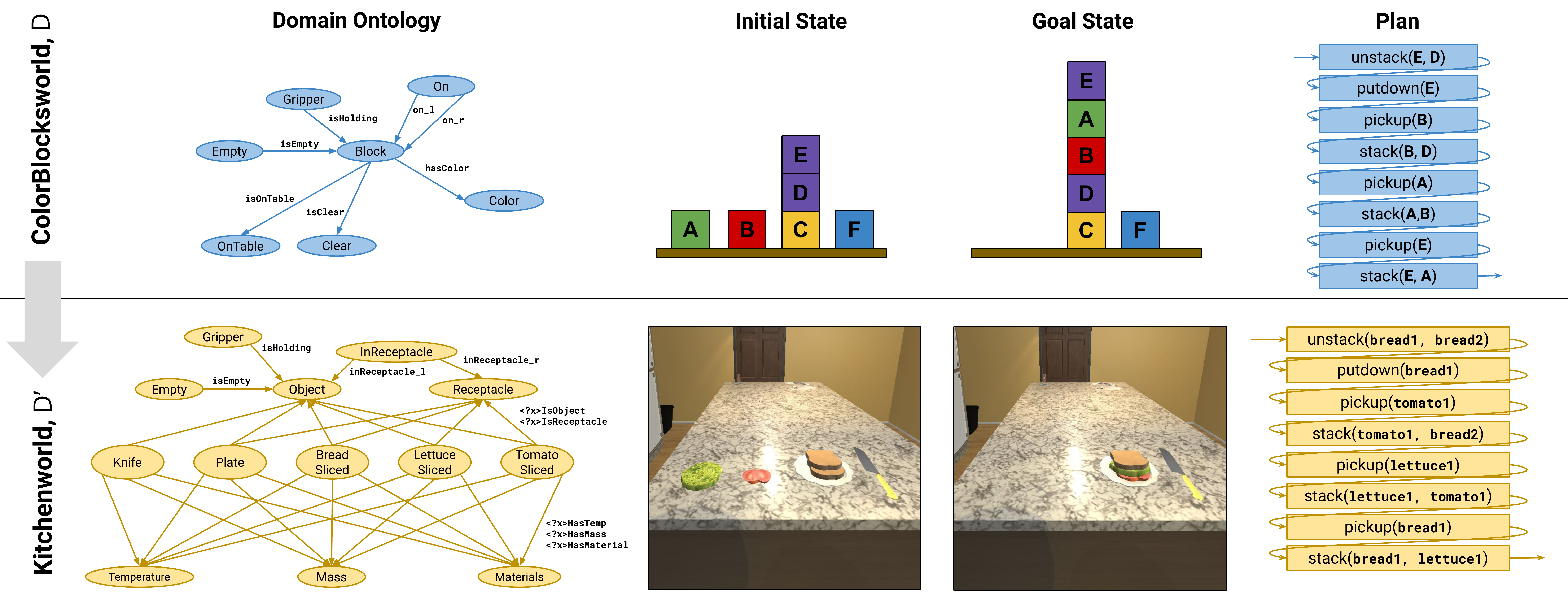}
  \caption{Illustration of plan transfer from the ColorBlocksworld domain to the Kitchenworld domain. The domain ontologies are drawn as directed graphs where the nodes of the graphs are the types and the directed edges of the graph are predicates. The source of the edge is the first argument of the predicate and the target of the edge is the second argument.}
  \label{fig:example}
\end{figure*}

Despite their differences, these domains share structural similarities. For example, both involve stacking objects to achieve goal arrangements, which suggest analogies between plans in the two domains. In this section, we formalize this analogy and use it to transfer plans in a principled manner. The domain ontologies and planning problems referenced in this section are illustrated in Figure \ref{fig:example}. For the implementation, we use the AlgebraicJulia framework \cite{AlgebraicJulia}, a Julia-based toolkit for category theory. 

\subsection{Domain specifications}
Since planning domain ontologies are user-defined, we begin by identifying the key ontological features for each domain. Both domains should be able to express predicates such as the existence of blocks, ingredients, and robot grippers. They should also be able to express facts about entities being on other entities. Lastly, both domains should have the ability to define attributes such as color, temperature, and mass. All of these concepts can be expressed using schema categories, which we will refer to as $\cat{D}$ and $\cat{D}^{\prime}$, for ColorBlocksworld and Kitchenworld, respectively.

\paragraph{ColorBlocksworld Ontology}

The ColorBlocksworld ontology contains types for concepts in the world, such as:

\begin{minted}{text}
  Block, Empty, Gripper
\end{minted}

It also contains types for attributes, such as:
\begin{minted}{text}
  Clear, OnTable, Color
\end{minted} 

\noindent It also contains predicates, such as:

\begin{minted}{text}
  isHolding::Hom(Gripper,Block)
\end{minted}

that can be used to express the fact that the \jul{Gripper} is holding a block \jul{Block}. We also include the predicate:

\begin{minted}{text}
  isEmpty::Hom(Empty,Block)
\end{minted}

which can be used to identify a \jul{Block} as representing an empty block. This gives the option for the gripper to hold nothing, making the gripper empty. 

Additionally, a \jul{Block} can have the following attributes:

\begin{minted}{text}
  hasColor::Attr(Block, Color)
\end{minted}

which describes the color of the block:
\begin{minted}{text}
  isClear::Attr(Block, Clear)
\end{minted}

which expresses whether or not a block is on top of it, and:

\begin{minted}{text}
  isOnTable::Attr(Block, OnTable)
\end{minted}

which expresses whether or not a block is on the table. States in ColorBlocksworld assign these attribute types to data primitives such as \jul{String} for \jul{Color} and \jul{Bool} for \jul{Clear} and \jul{OnTable}.

Lastly, there is a type, \jul{On}, that can be used to express relations between blocks. \jul{On} has predicates:

\begin{minted}{text}
  on_l::Hom(On, Block)
  on_r::Hom(On, Block)
\end{minted}

where \jul{on_l} maps to the block on top and \jul{on_r} maps to the block on the bottom. A full specification of $\cat{D}$ can be found in Appendix \ref{app:blocksworld}.

The actions of the ColorBlocksworld domain are modeled after the PDDL specification of Blocksworld \cite{pddl4j_blocksworld}. For example, the stack action, \jul{stack(top::Block, bottom::Block)}, puts the top block on the bottom block assuming the gripper is holding the block. The unstack action, \jul{unstack(top::Block, bottom::Block)}, removes the top block from the bottom block and keeps the block in the gripper. The precondition for both actions involve a conjunction of predicates that state that the top block is on (or not on) the bottom block and the gripper is empty (or not). The effect in both actions include a conjunction of predicates that express a complement of the precondition. In our representation language, the unstack action can be expressed using the following action span:

\begin{minted}[fontsize=\small]{text}
:unstack => @migration(OntBlocksworld, begin
  Pre => @join begin
    gripper::Gripper
    empty::Empty
    (block, underblock)::Block
    o::InOn
    inOn_l(o) == block
    inOn_r(o) == underblock
    isClear(block) == true
    isHolding(gripper) == isEmpty(empty)
  end
  Keep => @join begin
    gripper::Gripper
    (block, underblock)::Block
  end
  Eff => @join begin
    gripper::Gripper
    (block, underblock)::Block
    isClear(underblock) == true
    isHolding(gripper) == block
  end
end)
\end{minted}

The remaining actions of Blocksworld can be expressed in a similar way. 


\paragraph{Kitchenworld Ontology}

The Kitchenworld ontology was constructed by selecting a subset of types from the AI2-THOR world specification \cite{AiTHOR_object_types}, focusing on those necessary for the task of building a sandwich. While the full ontology could theoretically be represented, this subset was intentionally chosen for pedagogical clarity and to simplify the case study demonstration. However, the proposed method is not limited to such a small ontology—a larger and more varied ontology, including more utensils or liquids, could be incorporated without fundamental changes to the methodology.

Compared to ColorBlocksworld, Kitchenworld differs in its types and predicates.  Most notably, instead of blocks, there are objects, receptacles, ingredients, and utensils. Additionally, ingredients in Kitchenworld have attributes such as temperature, mass, and material instead of color. 

In this example, there exist types that represent ingredients, such as:

\begin{minted}{text}
  BreadSliced, LettuceSliced, TomatoSliced
\end{minted}

and utensils, such as:

\begin{minted}{text}
  Plate, Knife
\end{minted} 

All of these items are children types of an AI2-THOR \jul{Object}. This type hierarchy is expressed using predicates of the form:

\begin{minted}{text}
  plateIsObject::Hom(Plate, Object)
\end{minted}

which can be read as "Every plate is an object".

Receptacles are things for which other objects can be placed in or on top of. Some ingredient and utensil types be children types of \jul{Receptacle}. These predicates can be  expressed as:

\begin{minted}{text}
  plateIsReceptacle::Hom(Plate, Receptacle)
\end{minted}

In this ontology, bread slices, lettuce slices, tomato slices, and plates are receptacles and knives as not.

Each item in the kitchen can have the attributes---temperature, mass, and salient material according to \cite{AiTHOR_object_types}. These attributes are expressed as follows:

\begin{minted}{text}
  plateHasTemperature::Attr(Plate, Temperature)
  plateHasMass::Attr(Plate, Mass)
  plateHasMaterial::Attr(Plate, Material)
\end{minted}

Similar predicates can be expressed for each ingredient and utensil type. States in Kitchenworld assign these attribute types to data primitives such as \jul{Float64} for \jul{Temperature} and \jul{Mass} and \jul{String} for \jul{Material}.

The Kitchenworld domain shares some common features with the Blocksworld domain. For instance, it includes the types \jul{Empty} and \jul{Gripper}. It also includes their corresponding predicates that describe that the gripper is holding an item or is empty. It also has a type that mimics the \jul{On} type from ColorBlocksworld, namely \jul{InReceptacle}, which represents an object being in a receptacle. Similarly, this includes a left, \jul{inReceptacle_l}, and right, \jul{inReceptacle_r}, predicate that identifies which object is in which receptacle, respectively. A full specification of $\cat{D^{\prime}}$ can be found in Appendix \ref{app:kitchenworld}.

As for actions, there are many offered by the AI2-THOR specification, however, this approach adapts the actions from the ColorBlocksworld domain to be used in the Kitchenworld domain. This means that actions in the Kitchenworld domain do not need to be defined.

\subsection{Ontology Map}
The domain ontologies have common features, but also differ in a few ways. Formally, we can manually express an analogy that accounts for these observations using a ontology map, $F: \cat{D}^{\prime} \rightarrow \cat{D}$, particularly one that supports a conjunctive query data migration. Ontology maps must map target types to diagrams of source types and target predicates to maps between diagrams in a way that preserves compositions. In this example, we map each type and predicate in the Kitchenworld domain to diagrams of types and maps between them in the ColorBlocksworld domain.

First, there are obvious mappings for types and predicates with similar semantics, such as:

\begin{minted}[fontsize=\small]{text}
  # types
  Gripper => Gripper
  Empty => Empty
  # predicates
  isHolding => isHolding
  isEmpty => isEmpty
\end{minted}

Next, we focus our attention on the central types, namely \jul{Block}, \jul{Object}, \jul{Receptacle}. Here, we can leverage the intuition that a block can represent the most general notion of an item in Kitchenworld. In the case of Kitchenworld, both objects and receptacles can be interpreted as blocks in ColorBlocksworld. In addition, both domains have ways of expressing items being on top of other. Therefore, we can define the following component maps for $F$:

\begin{minted}[fontsize=\small]{text}
  # types
  Object => Block
  Receptacle => Block
  InReceptacle => On
  # predicates
  inReceptacle_l => on_l
  inReceptacle_r => on_r
\end{minted}

Recall that ColorBlocksworld provides us with information about block color. In our translation, we would like to map particular colored blocks to ingredients and utensils in Kitchenworld. For example, we would like to say that all purple blocks map to bread slices. We can express this assignment as:

\begin{minted}[fontsize=\small]{text}
  BreadSliced => @join begin
    block::Block
    C::Color
    (hc:block → C)::hasColor
    (c:block → C)::(block -> "purple")
  end
\end{minted}

The same can be done for every block color. In this case, we say that red blocks map to tomato slices, green blocks map to lettuce slices, yellow blocks map to plates, and blue blocks map to knives.

Lastly, we consider their attributes. Because there does not exist a notion of temperature, mass, and material of blocks in ColorBlocksworld, we must resort to mapping these attributes to primitive data types and default values for each child type. For example:

\begin{minted}[fontsize=\small]{text}
  Temperature => Float64
  Mass => Float64
  Material => String
\end{minted}

We can map each predicate of \jul{BreadSliced} to a function that assigns the following default values for every instance of BreadSliced, namely temperature as \jul{0.0}, mass as \jul{0.0}, and materials as \jul{"Unknown"}:

\begin{minted}[fontsize=\small]{text}
  breadSlicedHasTemperature => (x -> 0.0)
  breadSlicedHasMass => (x -> 0.0)
  breadSlicedHasMaterial => (x -> "Unknown")
\end{minted}

The same can be done for all ingredient and utensil types.

While it is mathematically possible to define functions that compute more informed values for these attributes, doing so introduces additional complexity. In particular, if these functions are multivariate and depend on other values in the state, they may introduce interdependencies that complicate the functoriality of the generated states. This, in turn, can lead to a plan transfer process that is not well-behaved.

Additionally, notice that the ColorBlocksworld attributes \jul{Clear} and \jul{OnTable} are not accounted for in this translation because there does not exist an analogous concept with the Kitchenworld domain. The data contained in those attributes are ignored which result in a lossy translation. In this scenario, this is an acceptable outcome, however, we acknowledge that in others this might not be. In which case, this would require that the domain ontologies be tweaked to ensure a lossless translation.

\begin{table*}[ht!]
  \centering
  \caption{Transferred initial state from ColorBlocksworld to Kitchenworld}
  \begin{tabular}{cc}
  \textbf{ColorBlocksworld ($\catSet{D}$)} & \textbf{Kitchenworld ($\catSet{D^{\prime}}$)} \\
  \hline
  \begin{minipage}[t]{0.8\columnwidth}
  \vspace{1mm}
  \begin{minted}[fontsize=\scriptsize]{text}
  # Initial state
  gripper::Gripper
  empty::Empty
  isHolding(gripper) == isEmpty(empty)
  (A, B, C, D, E, F)::Block
  hasColor(A) == "green"
  hasColor(B) == "red"
  hasColor(C) == "yellow"
  hasColor(D) == "purple"
  hasColor(E) == "purple"
  hasColor(F) == "blue"
  (x1, x2)::On
  on_l(x1) == E 
  on_r(x1) == D
  on_l(x2) == D 
  on_r(x2) == C
  isOnTable(A) == true; isClear(A) == true
  isOnTable(B) == true; isClear(B) == true
  isOnTable(C) == true; isClear(C) == false
  isOnTable(D) == false; isClear(D) == false
  isOnTable(E) == false; isClear(E) == true
  isOnTable(F) == true; isClear(F) == true
  \end{minted}
  \vspace{2mm}
  \end{minipage}
  &
  \begin{minipage}[t]{0.8\columnwidth}
  \vspace{1mm}
  \begin{minted}[fontsize=\scriptsize]{text}
  # Initial state (after transfer)
  gripper::Gripper
  empty::Empty
  isHolding(gripper) == isEmpty(empty)
  A::LettuceSliced
  B::TomatoSliced
  C::Plate
  D::BreadSliced
  E::BreadSliced
  F::Knife
  (x1, x2)::InReceptacle
  inReceptacle_l(x1) == BreadSlicedIsObject(E) 
  inReceptacle_r(x1) == BreadSlicedIsReceptacle(D)
  inReceptacle_l(x2) == BreadSlicedIsObject(D)
  inReceptacle_r(x2) == PlateIsReceptacle(C)
  lettuceHasTemperature(A) == 0.0
  lettuceHasMass(A) == 0.0
  lettuceHasMaterial(A) == "Unknown"
  tomatoHasTemperature(B) == 0.0
  tomatoHasMass(B) == 0.0
  tomatoHasMaterial(B) == "Unknown"
  ...

  \end{minted}
  \vspace{2mm}
  \end{minipage}
  \\
  \hline
  \end{tabular}
  \label{tab:code}
\end{table*}

\subsection{Plan Transfer from ColorBlocksworld to Kitchenworld}
Provided an ontology map, we can now transfer a plan based on a specific problem in Blocksworld.

\subsubsection{ColorBlocksworld problem}
Let us consider a problem in ColorBlocksworld where we have the initial block arrangement shown in Figure \ref{fig:example}. Here, there are six blocks, uniquely identified by letters \jul{{A,B,C,D,E,F}}, arranged on the table. Every block can be one of the following colors: \jul{"red"}, \jul{"green"}, \jul{"blue"}, \jul{"purple"}, \jul{"yellow"}. As an example, notice there are two blocks, \jul{E} and \jul{D}, that are the color purple. The goal is to arrange the blocks into a stack such that the colors of the blocks are \jul{purple, green, red, purple, yellow} from top to bottom. We can specify the initial and goal state as functors in $\catSet{D}$ by specifying a map from each type in $\cat{D}$ to an element in $\cat{Set}$. As an example, the minimum specification of the initial state in ColorBlocksworld is shown in Table \ref{tab:code} (left). The goal state can be specified using the same principles.

A plan for this problem could consist of the following sequence of grounded actions:

\vspace{0.1cm}

\begin{minted}[fontsize=\small]{text}
unstack(E, D); putdown(E); 
pickup(B); stack(B, D); pickup(A); 
stack(A, B); pickup(E); 
stack(E, A)
\end{minted}

\vspace{0.1cm}

\subsubsection{Transfer}
To transfer this plan, each action and its match must be transferred. Consider a plan where the first action is to unstack block \jul{E} from \jul{D}. In this scenario, the match for the unstack action grounds the predicate \jul{block} to \jul{E} and \jul{underblock} to \jul{D}. To understand how the plan transfer functor $\Delta_F$ impacts this plan, we examine how $\Delta_F$ acts on the initial action, \jul{unstack(E, D)}, and its match. The functor $\Delta_F$ transforms the planning domain by sending $X$ to $X \circ F$, in the simplest case, which has the signature $\cat{D}^{\prime} \rightarrow \cat{D} \rightarrow \cat{Set}$. This means that for every type in $D^{\prime}$, we can follow the functions to its corresponding set in $\cat{Set}$. This process produces the transferred state, ensuring that the plan is consistent within the new domain.

Table \ref{tab:code} shows the transferred initial state from ColorBlocksworld to Kitchenworld. While the instance names, such as \jul{A, B, gripper}, remain unchanged, they are now considered instances of types in Kitchenworld. The $\mathrm{Keep}$ and $\mathrm{Eff}$ components of the action undergo a similar transfer. Finally, the match morphisms are also transferred. In summary, the transfer translates the types from ColorBlocksworld to Kitchenworld while preserving the semantics of the predicates. All remaining actions and intermediate states of the plan can be transferred in the same way, thus, creating a corresponding plan in Kitchenworld:


\begin{minted}[fontsize=\small]{text}
unstack(bread1, bread2); putdown(bread1); 
pickup(tomato); stack(tomato, bread2); 
pickup(lettuce); stack(lettuce, tomato); 
pickup(bread1); stack(bread1, lettuce);
\end{minted}


In this paper, the instance names have been adjusted to adopt contextually relevant names for Kitchenworld, such as \jul{bread1} and \jul{bread2}, to align with the domain's semantics. However, in the underlying code, these instances are represented by different symbols, which are mapped appropriately according to the transfer process.

\subsubsection{Generated actions}
Notice that Blocksworld-derived actions are abstract and domain-agnostic, allowing them to generalize across different environments and support high-level plan reasoning. In contrast, AI2-THOR actions \cite{ai2thor_interactive_physics} are tightly coupled to the physics-based simulation engine, meaning that their execution depends on low-level constraints, such as object interaction mechanics.

For example, while the actions \jul{putdown} and \jul{pickup} have direct counterparts in AI2-THOR’s available actions, higher-level relational actions like \jul{stack} and \jul{unstack} do not exist in AI2-THOR’s action models. This absence reflects AI2-THOR’s emphasis on physics-based interactions, where actions primarily operate at the level of individual object states rather than explicitly encoding relationships between objects. This suggests an opportunity for AI2-THOR to extend its action models by incorporating relational actions that capture object interactions. Notably, these additions could be adopted from the Blocksworld’s transferred actions.


\section{Discussion}

The proposed approach offers a principled method for transferring task plans from one planning context to another. We demonstrated its application by transferring plans from a simple domain, ColorBlocksworld, to a more complex domain, Kitchenworld. In this section, we discuss general patterns for using this approach, its integration into a planning architecture, and considerations when defining ontology maps.

\subsection{Patterns of Use for Functorial Plan Transfer}
\label{sec:patterns}

\begin{figure}[h!]
  \centering
  \includegraphics[width=\columnwidth]{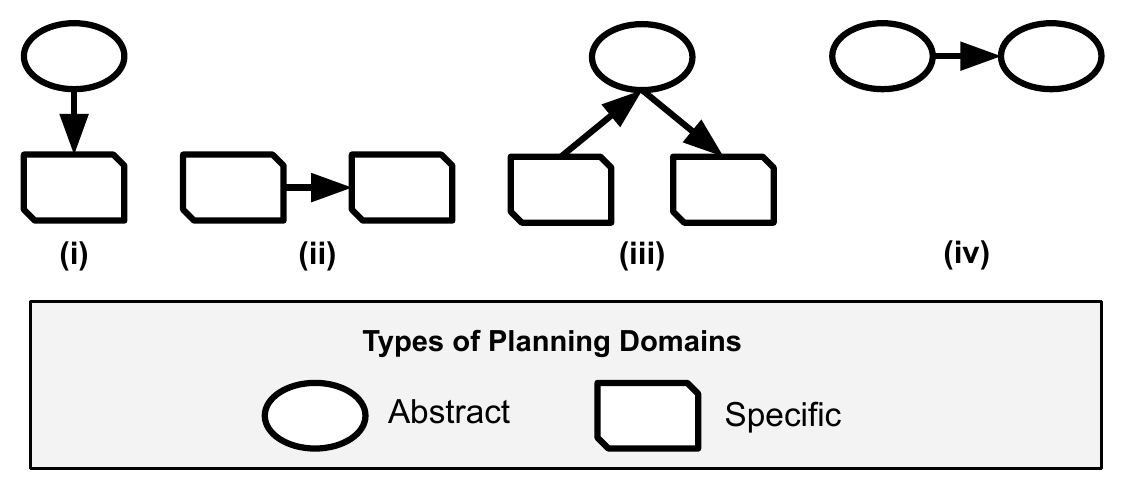}
  \caption{Schematic of the four patterns of functorial plan transfer: (i) abstract-to-specific, (ii) specific-to-specific, (iii) specific-to-abstract-to-specific, and (iv) abstract-to-abstract.}
  \label{fig:transferpatterns}
\end{figure}

As demonstrated by the case study in Section \ref{sec:case}, planning domains can differ in their level of detail and generalization. An \textit{abstract} domain represents a simplified or generalized environment with fewer and well-defined concepts, such as Blocksworld. In contrast, a \textit{specific} domain is a detailed and complex environment with many domain-specific attributes, such as Kitchenworld. Based on this distinction, we have identified four patterns of use: (i) abstract-to-specific, (ii) specific-to-specific, (iii) specific-to-abstract-to-specific, and (iv) abstract-to-abstract. Each pattern serves a unique purpose in robotic task planning and highlights the versatility of the approach. A schematic of each pattern is shown in Figure \ref{fig:transferpatterns}.

\subsubsection{Abstract-to-Specific}
This pattern is exemplified by the Blocksworld-to-Kitchenworld example presented in this paper. In this approach, a generic plan developed in an abstract domain is transferred to a specific domain. The advantage of this pattern lies in its ability to reapply generalized behaviors to specific contexts without compromising the integrity of the original behavior.

An extended application of this pattern involves migrating a plan from a simulation environment—where domain concepts are fewer and well-defined—to a real-world environment \cite{Pitkevich2024}, which is richer but less structured. For example, reasoning in a real-world kitchen might require capturing additional relationships, such as the relative spatial alignment of objects on a cluttered countertop. This pattern enables structured actions from a simplified domain to be effectively used in more complex and dynamic settings, while offering computational advantages by reasoning in the abstract domain.

\subsubsection{Specific-to-Specific}
In this pattern, a plan is directly transferred from one specific domain to another. For instance, a plan developed for tending a shopfloor machine might be adapted to a similar domain-specific task, such as managing an industrial oven for food production—both requiring managing items being processed in a machine. The advantage of this approach lies in its ability to replicate context-specific behaviors that may not be easily generalized or articulated in an abstract domain.

This pattern closely aligns with ongoing research in skill generation \cite{Wang2023,Garcia2015,VieiraDaSilva2023,Li2023}, where plans are designed for highly specialized tasks. By transferring such plans directly between specific domains, it becomes possible to preserve nuanced, context-dependent behaviors that could otherwise be lost during abstraction.

\subsubsection{Specific-to-Abstract-to-Specific}
This pattern involves transferring a plan from one specific domain to an abstract domain, distilling its essential structure, and then applying it to a different specific domain. For example, Factory A’s machine-tending plan could be generalized into an abstract template, which is subsequently tailored to fit the operational needs of Factory B.

The primary benefit of this pattern is its ability to generalize and distill the key elements of a complex plan, making it applicable to a wider range of scenarios. This is particularly valuable in privacy-sensitive collaborations \cite{Zhang2020SPAR,Alsayegh2023,Demarinis2019,Park2020}, where manufacturers may wish to adhere to a shared abstract template while safeguarding proprietary details. By relying on the abstract-specific transfer, they can impose generic behavior in different contexts while maintaining confidentiality.

\subsubsection{Abstract-to-Abstract}
This pattern involves transferring plans between two abstract domains. For example, a plan from the Tower of Hanoi domain \cite{Havur2013} could be adapted for the Blocksworld domain. This pattern introduces a novel concept in AI planning: reframing a planning problem from one canonical abstract domain to another to potentially gain computational or reasoning advantages. For instance, a planner might exploit constraints related to relative size and position in the Tower of Hanoi to identify an intermediate subgoal, then continue planning in Blocksworld.

This pattern holds significant potential for algorithmic reframing of abstract search problems in robotics applications. In particular, it could prove useful in scenarios where the structured constraints of a Tower of Hanoi problem offer greater efficiency compared to the less structured nature of a Blocksworld problem. Exploring how such reframing can improve planning efficiency and problem-solving capabilities is a promising direction for future research.

\subsection{Integration into a Planning Architecture}
The described transfer use cases can be integrated into a planning architecture similar to case-based planning \cite{Jung2007}. In this framework, source plans are retrieved from a historical knowledge base, transferred, and adapted to suit specific use cases. The transferred plan acts as a skeleton or backbone for the new plan, ensuring that analogous behavior is preserved as long as the sequence of actions is maintained. The approach may require plan repair, particularly when actions in the transferred plan must be decomposed into more granular actions. For instance, translating a high-level plan or program into a low-level robot programming language might necessitate finer action descriptions \cite{Aguinaldo2021}. Developing a mathematically rigorous framework to address such decompositions is an area of ongoing research.

\subsection{Designing an Ontology Map}


The success of all the aforementioned patterns, in Section \ref{sec:patterns}, is contingent on the ability to establish a valid ontology map between the source and target domains. The design of the onotology map is critical, as it directly influences the semantics and feasibility of the transferred plan.

The structure and mappings of an ontology map dictate whether a meaningful and correct analogy between domains can be established. For instance, consider an inappropriate translation for the Kitchenworld and Blocksworld domains that involves mapping the \jul{Gripper} from Kitchenworld to \jul{Block} in Blocksworld--expressing a nonsensical analogy. Validating this mapping against the properties of a functor would reveal that no proper ontology map can be defined, thus preventing plan transfer.

Alternatively, a semantically flawed but technically valid translation might involve swapping the mappings of \jul{inReceptacle_l} to \jul{on_r} and \jul{inReceptacle_r} to \jul{on_l}. This reversal would invert the stacking order in Kitchenworld, leading to a semantically undesirable plan.

\subsubsection{Failure to Define an Ontology Map}

The inability to define an ontology map indicates the absence of a proper analogy between domains, which directly prevents reliable plan transfer. Fortunately, such failures are identified during the design stage of the translation process, allowing the issue to be addressed before runtime.

It is, however, important to note that the failure to define an ontology map may sometimes stem from structural mismatches in the domain ontologies rather than an inherent inability to transfer plans. Careful design of both source and target ontologies is therefore critical. Ontology development requires deliberate decisions to represent concepts in ways that facilitate alignment.

For example, consider a choice to make \jul{Receptacle} a subtype of \jul{Object} (\jul{Receptacle} $\rightarrow$ \jul{Object}) in Kitchenworld. While valid, this design choice would prohibit an ontology map that assigns both \jul{Object} and \jul{Receptacle} to \jul{Block} in Blocksworld, as no predicate morphism exists from \jul{Block} to \jul{Block} in Blocksworld. By contrast, separating \jul{Receptacle} and \jul{Object} in Kitchenworld better aligns with the \jul{Block} schema in Blocksworld, allowing an ontology map to be defined. This example underscores the importance of ontology design in ensuring successful plan transfer.

\subsubsection{Addressing Non-One-to-One Ontology Mappings}

Ontological misalignments between domains are a common challenge in practice, emphasizing the importance of conjunctive query migrations for capturing nuanced relationships. Structural features such as the size of an ontology—defined by the number of types ($|\cdot|$)---or the overlap of aligned content can significantly influence the success of a plan transfer. These features give rise to the following scenarios:

\paragraph{$|\text{Target}| > |\text{Source}|$: Stratifying State Data}

As described in the case study in Section \ref{sec:case}, a task plan originating in Blocksworld was successfully transferred to Kitchenworld. In this example, the Kitchenworld domain ontology is larger than the Blocksworld ontology, requiring multiple types in the target domain to be mapped to a single type, \jul{Block}, in the source domain. This mapping stratifies block instances, aligning them with different types of ingredients in Kitchenworld. The conjunctive query migration functor plays the crucial role of identifying and aligning these stratifications, ensuring that instances of the Kitchenworld ontology are populated correctly.

\paragraph{$|\text{Target}| < |\text{Source}|$: Merging State Data}

In scenarios where a plan is transferred from Kitchenworld to Blocksworld, the target domain ontology is smaller than the source ontology. This results in the loss of distinctions, such as different types of ingredients, as all objects are reduced to a single \jul{Block} type. In such cases, the designer must carefully evaluate whether this simplification affects the validity or applicability of the transferred plan, particularly when domain-specific attributes are critical for task success.

\paragraph{Target $\cap$ Source: Information Loss}

When the source and target ontologies overlap but have non-identical structures, some data loss is inevitable. This loss may be acceptable in certain contexts but detrimental in others. For instance, attributes like \jul{isClear}, \jul{onTable}, \jul{mass}, and \jul{temperature} might be omitted unless explicitly accommodated in the target ontology. Acceptable loss occurs when the goal is to transfer the sequence of stacking actions without leveraging detailed attributes, as such omissions may not affect the transfer’s success. Conversely, detrimental loss arises when plans rely on specific attribute values, such as ensuring a weight limit of 100 lb for stacking. Defaulting these values to placeholders, such as \jul{weight = 0}, can invalidate the plan. Addressing these issues requires additional validation steps, such as populating missing values or adapting the transferred plan. Developing a mathematically robust framework for handling missing attributes remains a promising area for future research.

\section{Proposed Metrics for Evaluating Plan Transfer Approaches}
\label{sec:metrics}

This paper introduces a novel approach for symbolically transferring task plans across planning domains, which presents challenges for direct comparison with existing methods due to the lack of similarly established techniques. However, benchmarks are essential for evaluating the method's effectiveness, practicality, and explainability. They also serve as a foundation for systematic assessment and comparison with future advancements. In this section, we propose key benchmarks for evaluating symbolic plan transfer.

\subsection{Metric 1: Percentage of Goal Satisfaction}
A fundamental benchmark is goal satisfaction, which measures the extent to which the transferred plan satisfies the desired goal state in the target domain. This metric evaluates the alignment between the source domain’s goal and the final state in the target domain. The transferred goal is decomposed into individual statements, and the percentage of statements satisfied in the final state is calculated. This approach is conceptually similar to goal achievement metrics used in AI planning, such as those employed in the International Planning Competition (IPC), where goal satisfaction is a key evaluation criterion \cite{Taitler2024}.

For example, in the case study presented in Section \ref{sec:case}, this involves verifying that goal statements—such as the stacking order of ingredients—are fully satisfied.

In the general case of delta data migrations, this metric consistently achieves 100\% goal satisfaction when the ontology map is well-defined (functorial) and the transferred goal is semantically compatible. In contrast, conjunctive query migrations may yield varied results depending on the mappings and constraints. The procedure for validating goal satisfaction in both cases is described in Section \ref{sec:validate}. This benchmark provides a clear measure of how effectively the transfer method preserves the intended goal state.

\subsection{Metric 2: Explainability of the Plan Transfer Process}
Explainability ensures that human operators can understand and validate the rationale behind the plan transfer. This benchmark focuses on two key aspects:

\begin{itemize}
\item \textbf{Comprehension of User-Defined Mappings:} Assessing how well users understand the mappings used to transfer plans between domains, such as the ontology map in this approach.
\item \textbf{Comparison of Source and Target Plans:} Determining whether the transferred plan is a clear and logical analog of the source plan.
\end{itemize}

In our approach, the ontological mappings between planning domains would be presented alongside the source and transferred plans. Participants would evaluate their ease of understanding using methods such as Likert scale ratings or confidence levels. Additionally, targeted questions (e.g., "Why do you think the transferred state is correct based on this mapping?") could be asked to offer insights into participants' understanding of the process. This benchmark provides a clear measure of the ease at which a human operator can discern the validity of the transferred plan.

\subsection{Metric 3: Ease of Translation to Motion Plans}
Symbolic plans must effectively bridge the gap to executable robot commands. This benchmark evaluates the transition process by measuring factors such as the time, computational resources, and error rates involved in converting symbolic plans into motion trajectories. The evaluation should be normalized based on the translation method used, such as direct mapping of symbolic actions to motion primitives, the use of intermediate task representations, or learning-based approaches for motion plan generation \cite{Garrett2021,Dantam2018,Guo2023}.

For instance, in the case study presented in Section \ref{sec:case}, this benchmark would assess the effort required to generate robot trajectories for symbolically represented actions like picking, placing, and stacking ingredients for a real-world kitchen. This benchmark provides a clear measure of efficiency in the translation from high-level to low-level robot commands which influences the practical utility of symbolic plan transfer.

\subsection{Metric 4: Physical Feasibility of the Transferred Plan}
The physical feasibility of the transferred plan is a critical benchmark for practical applications. This metric evaluates whether the transferred plan can be successfully executed in a real-world or simulated environment.

For example, in the case study presented in Section \ref{sec:case}, this involves deploying the transferred plan on a robot to assemble the sandwich in a physical or simulated kitchen and observing its performance. This benchmark provides a clear measure of viability of transferred plans in real-world environments.

\subsection{Metric 5: Efficiency of Plan Transfer vs. Replanning}
An essential benchmark for evaluating the utility and effectiveness of plan transfer is its efficiency compared to planning from scratch. This metric assesses whether transferring a plan offers measurable advantages in terms of time, computational resources, or solution quality compared to generating a new plan entirely. Key metrics for evaluating the efficiency of plan transfer include: 

\begin{enumerate}
\item \textbf{Planning Time and Computational Effort:} This metric compares the time required to transfer a plan versus plan from scratch. It includes computational factors such as CPU time, memory usage, and algorithmic complexity.
\item \textbf{Plan Quality:}
Plan quality is determined by comparing the number of steps in the transferred plan to those in a newly generated plan. Additionally, the metric considers the edit distance between a plan created directly in the target domain and one obtained through plan transfer. This measure, also known as \textit{plan stability} \cite{Fox2006}, indicates how much the transferred plan deviates from an independently generated one.
\end{enumerate}

In the case study presented in Section \ref{sec:case}, this metric would compare the computational effort and quality attained from transferring a Blocksworld plan to the Kitchenworld domain versus generating a new plan from scratch in Kitchenworld.

This metric is inspired by studies in transfer learning and classical AI planning, which examine the benefits of transferring knowledge versus learning from scratch \cite{Taylor2009} and the reuse of plan fragments to accelerate planning \cite{Bergmann1996}.


\section{Related Works}

This paper presents an approach to symbolic task plan transfer based well-structured ontology maps. As such, we have identified a few key areas of related work: skill transfer in robotics, domain mapping and plan migration, and large language model (LLM)-based approaches in robot planning.

\subsection{Skill Transfer in Robotics}
Skill transfer in robotics focuses on translating skills, or plans, from one scenario to another to avoid costly replanning and domain design. Planning scenarios can vary based on the specific task or environment \cite{Jaquier2023}. Plan transfer approaches can be broadly classified into two categories: source-to-target transfer (direct translation) and source-to-abstract-to-target transfer (task generalization). 

\paragraph{Direct Translation}
Direct translation methods map source plans directly to target domains using data-driven techniques such as implicit behavior cloning \cite{Zhang2023}, multi-task reinforcement learning frameworks \cite{Bajpai2018, Lee2021, Sun2023, Men2023, Zhao2022}, and progressive neural networks \cite{Meng2024}. While effective, these methods are often domain-specific point solutions and lack generalizability across all plans in a domain.

\paragraph{Task Generalization}
Task generalization focuses on identifying and utilizing minimal representations of tasks to enable transfer \cite{Li2023,Jin2025,Hao2022,Jin2024}. Abstracted plans are cached and reused with modifications when similar tasks arise. Case-based planning in classical AI planning adopts a similar approach \cite{Bergmann1996, Hanks1995} but is limited to adapting plans within the same domain and has not been explored in cross-domain transfer scenarios.

\subsection{Domain Mapping and Plan Migration}
Domain mapping efforts focus on translating tasks across domains \cite{Elimelech2023SPAR1, Elimelech2023SPAR2, Men2023, Morere2019, Heuss2023}. For example, Heuss et al. \cite{Heuss2023} proposed mapping from abstract pick-and-place domains to specific kitting domains but relied on replanning rather than migrating existing plans. Similarly, Elimelech et al. \cite{Elimelech2023SPAR1, Elimelech2023SPAR2} introduced projections, a form of abstracted states that retain only salient attributes necessary for execution. However, these methods often neglect the role of ontological structures beyond attributes, which are crucial for robust domain mapping.

\subsection{LLM-Based Approaches in Robot Task Planning}
This work fundamentally differs from LLM-based robot task planning frameworks by focusing on formal, structured planning representations rather than leveraging generative models for transfer or state abstraction. Many LLM-based methods act as cognitive tools to interpret human intent or dynamically refine plans through human assistance \cite{GuangLi2023, Zhen2024, BastosdaSilva2024, Yang2024, Izquierdo-Badiola2024, Shin2024}. To date, LLMs are neither the sole nor the primary engine for planning or plan transfer. As a result, they do not fully address challenges of expressivity and semantic consistency during planning and plan transfer.

In contrast, the proposed method uses a rigorous category-theoretic framework to guarantee that task plans adhere to ontological structures. Unlike LLMs, which often augment classical planners \cite{Pallagani2024, Ding2023, Ding2023a}, this framework directly integrates expressivity and structure into the planning and plan transfer process, avoiding reliance on in-situ human or agent intervention.

\section{Future Work}
This paper lays the foundation for novel approaches to task plan transfer. Future work includes developing a benchmarking framework and datasets, empirical scaling studies, and mathematical analysis of conjunctive query migrations.

\paragraph{Benchmarking Framework and Datasets}
Developing a standardized benchmarking framework and curated datasets is essential for evaluating and comparing task plan transfer methods. Such a framework would include implementations of the benchmark metrics described in Section \ref{sec:metrics} and datasets spanning diverse domains, such as logistics, assembly tasks, service robotics, and canonical planning domains. These datasets would incorporate predefined ontologies, source plans, and real-world constraints to enable systematic evaluation.

\paragraph{Empirical Scaling Studies}
Understanding how this approach scales with increasing complexity is crucial. Future work will involve empirical studies to evaluate the scalability of our plan transfer method in scenarios with larger ontologies and state data, such as those derived from the KnowRob \cite{Tenorth2017} and Factory of the Future (FoF) \cite{Schafer2021} ontologies. These studies will also investigate the impact of non-trivial ontology maps on success rates and computational efficiency, with a focus on comparisons to replanning methods.

\paragraph{Mathematical Analysis of Conjunctive Query Migrations}
Conjunctive query migrations use ontology maps to define complex correspondences between types in one domain and diagrams of types in another. A deeper mathematical analysis of the conditions under which these migrations are functorial—or, more simply, guaranteed to produce valid plans—is an important area of research. Such an analysis will strengthen the theoretical foundations of this approach and enhance the expressivity and utility of a functorial plan transfer approach.

\section{Conclusion}
In conclusion, this paper highlights the necessity of advancing methods in plan transfer to make robot task plans applicable across various domains. By introducing a formalism based on functorial data migration and leveraging the principles of category theory, we provide a structured framework for generalizing task plans and enabling their reuse in new settings without extensive replanning. We demonstrated the feasibility of this approach through a case study that successfully transferred a plan from the ColorBlocksworld domain to one compatible with the AI2-THOR Kitchen simulation environment. Additionally, we identified general patterns of use for this method which include applications to privacy-preserving task planning and skill generation. Future work will focus on developing standardized benchmarking frameworks and datasets, empirically evaluating the scalability of the approach with large ontologies and state data, and conducting mathematical analyses of conjunctive query migrations to support more sophisticated plan transfer techniques.


\bibliographystyle{IEEEtran}
\bibliography{ref}

\vspace{-1cm}

\begin{IEEEbiography}[{\includegraphics[width=1in,height=1.25in,clip,keepaspectratio]{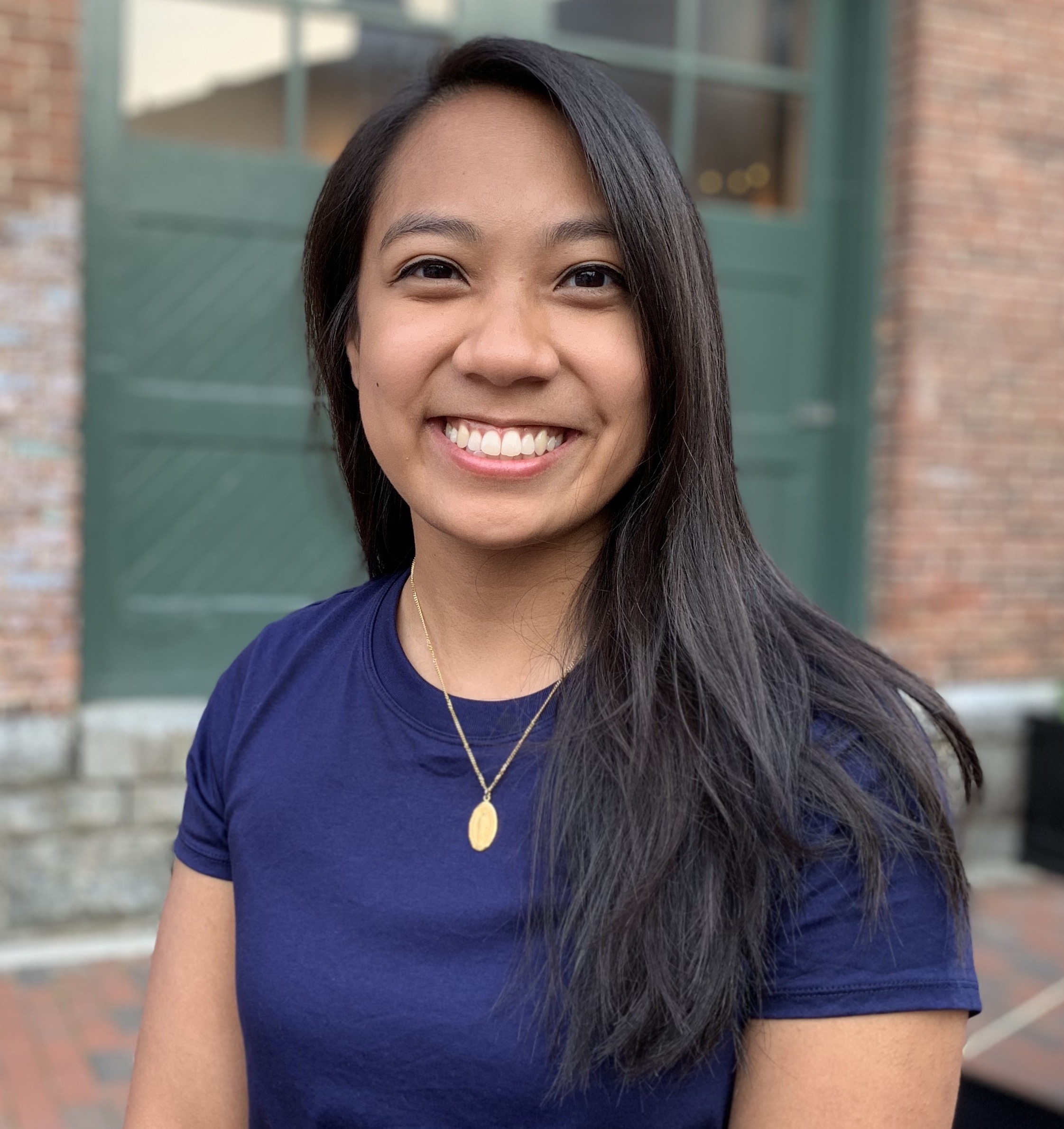}}]{Angeline Aguinaldo} (Member IEEE) is a senior research staff member at the Johns Hopkins University Applied Physics Laboratory. Her research focuses on applications of category theory to robotics, knowledge representation, and AI planning. She received a Ph.D. in Computer Science from the University of Maryland, College Park. She received a Master of Science degree in Electrical Engineering and a Bachelor of Science degree from Drexel University.
\end{IEEEbiography}

\vspace{-1cm}

\begin{IEEEbiography}[{\includegraphics[width=1in,height=1.25in,clip,keepaspectratio]{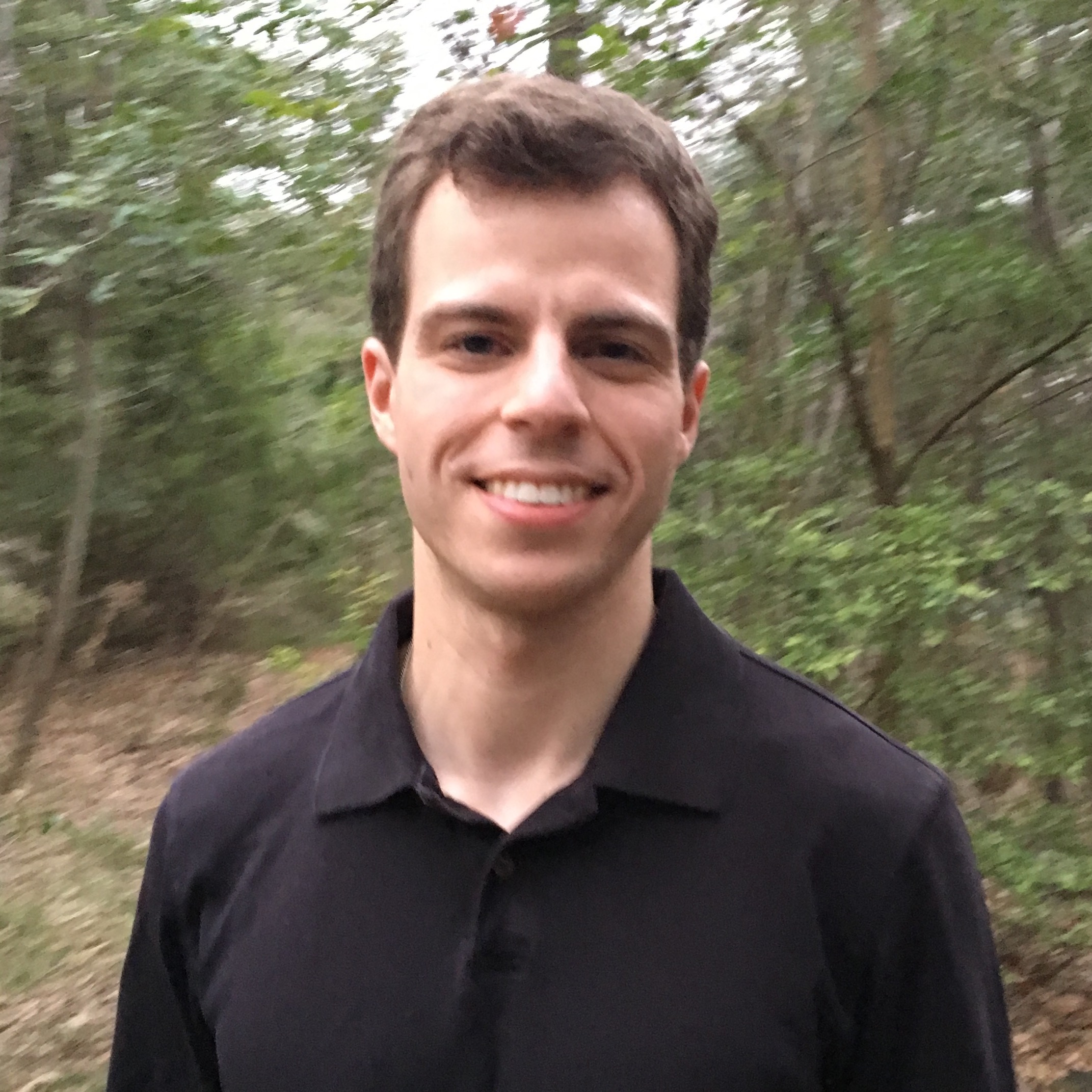}}]{Evan Patterson} is a Research Scientist at Topos Institute in Berkeley, California. His research focuses on category theory and applications thereof to scientific modeling and software systems. He received a Ph.D. in Statistics from Stanford University and a Bachelor of Science degree in Mathematics and Physics from Caltech.
\end{IEEEbiography}

\vspace{-1cm}

\begin{IEEEbiography}[{\includegraphics[width=1in,height=1.25in,clip,keepaspectratio]{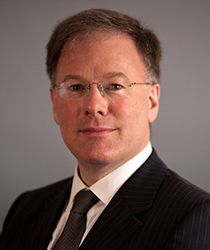}}]{William Regli} (Fellow IEEE) is a Professor of Computer Science at the University of Maryland at College Park. In addition to the University of Maryland, he has held positions with the Defense Advanced Research Projects Agency (DARPA), Lockheed Martin, AT\&T, and Drexel University. His research focus on applications of artificial intelligence to problems in engineering, manufacturing, communications, and other domains. Regli has published more than 250 technical articles and produced five U.S. Patents in the area of 3D CAD search. Dr. Regli holds a Ph.D. in Computer Science from the University of Maryland at College Park and Bachelor of Science degree in Mathematics from Saint Joseph's University. He is an elected Senior Member of both the ACM and AAAI; and a Fellow of AAAS and IEEE for his “contributions to 3D search, design repositories and intelligent manufacturing".
\end{IEEEbiography}




\newpage 
\appendices

\section{Proof of Valid Plan using Delta Data Migration}
\label{sec:proof} 

First, consider the final ground action in the source domain plan $\sigma_{\mathbf{D}}$. A match morphism, $m_n: \mathrm{Pre} \rightarrow Y_{(n-1)}$, consists of the object, $\mathrm{Pre}$, of the final action, and object $Y_{(n-1)}$, where $n$ is the total number of actions in the plan. Transferring this match morphism transfers its domain and codomain objects with it, which produces the morphism: 

\vspace{-0.2cm}

\[
\Delta_F(m_n) : \Delta_F(\mathrm{Pre}) \rightarrow \Delta_F(Y_{(n-1)})
\] 

\noindent and the second-to-last state $\Delta_F(Y_{(n-1)})$. Computing the DPO for this grounded action yields the final state in the target domain, which we denote by $Y^{\prime}_{n}$.

To verify that the goal has been satisfied in the target domain, it is necessary to check whether there exists a monomorphism in the target domain, namely:

\vspace{-0.2cm}

\[
g^{\prime}: \Delta_F(G) \hookrightarrow Y^{\prime}_{n},
\]

\noindent where $G$ is the goal state in the source domain. Given that the source plan satisfied the original goal, there must exist a monomorphism in the source domain:

\vspace{-0.2cm}

\[
g: G \hookrightarrow Y_{n}.
\]

\noindent Since delta data migrations preserve pushouts, we can conclude that:

\vspace{-0.2cm}

\[
\Delta_F(Y_{n}) = Y^{\prime}_{n}.
\]

\noindent By functoriality, the image of the monomorphism $g: G \hookrightarrow Y_{n}$ under $\Delta_F$ produces the monomorphism: 

\vspace{-0.2cm}

\[
\tilde{g}: \Delta_F(G) \hookrightarrow \Delta_F(Y_{n}).
\]

\noindent Since $\Delta_F(Y_{n}) = Y^{\prime}_{n}$, it follows that: 

\vspace{-0.2cm}

\[
\tilde{g}: \Delta_F(G) \hookrightarrow Y^{\prime}_{n}.
\]

Thus, we have shown that $\tilde{g} = g^{\prime}$ implying that there exists a monomorphism from the transferred goal $\Delta_F(G)$ to the final state $Y^{\prime}_{n}$ in the target domain. The fact that delta data migrations preserve pushouts ensures that this result holds in the general case, implying that all transferred plans between the source and target domains remain valid.

\section{Schema Categories}

\subsection{ColorBlocksworld}
\label{app:blocksworld}

A specification of the schema category for the Color Blocksworld domain ontology using AlgebraicJulia syntax.
\begin{minted}[fontsize=\footnotesize]{text}
@present OntBlocksworld(FreeSchema) begin
  # Parent Types
  Block::Ob
  Empty::Ob
  Gripper::Ob

  # Attributes
  Color::AttrType
  Clear::AttrType
  OnTable::AttrType

  # Relations 
  On::Ob
  on_l::Hom(InOn, Block)
  on_r::Hom(InOn, Block)

  # Robot
  isEmpty::Hom(Empty, Block)
  isHolding::Hom(Gripper, Block)

  # Blocks
  hasColor::Attr(Block, Color)
  isClear::Attr(Block, Clear)
  isOnTable::Attr(Block, OnTable)
end
\end{minted}

\subsection{Kitchenworld}
\label{app:kitchenworld}

A specification of the schema category for the Kitchenworld domain ontology using AlgebraicJulia syntax.

\begin{minted}[fontsize=\footnotesize]{text}
@present KitchenWorld(FreeSchema) begin
  # Parent Types
  Object::Ob
  Receptacle::Ob
  Empty::Ob
  Gripper::Ob

  # Robot
  isHolding::Hom(Gripper, Object)
  isEmpty::Hom(Empty, Object)

  # Attributes
  Temperature::AttrType
  Mass::AttrType
  Material::AttrType

  # Relations
  InReceptacle::Ob
  inReceptacle_l::Hom(InReceptacle, Object)
  inReceptacle_r::Hom(InReceptacle, Receptacle)

  # Ingredients
  BreadSliced::Ob
  breadSlicedIsObject::Hom(BreadSliced, Object)
  breadSlicedIsReceptacle::Hom(
    BreadSliced, Receptacle)
  breadSlicedHasTemperature::Attr(
    BreadSliced, Temperature)
  breadSlicedHasMass::Attr(BreadSliced, Mass)
  breadSlicedHasMaterial::Attr(
    BreadSliced, Material)

  LettuceSliced::Ob
  lettuceSlicedIsObject::Hom(
    LettuceSliced, Object)
  lettuceSlicedIsReceptacle::Hom(
    LettuceSliced, Receptacle)
  lettuceSlicedHasTemperature::Attr(
    LettuceSliced, Temperature)
  lettuceSlicedHasMass::Attr(
    LettuceSliced, Mass)
  lettuceSlicedHasMaterial::Attr(
    LettuceSliced, Material)

  TomatoSliced::Ob
  tomatoSlicedIsObject::Hom(
    TomatoSliced, Object)
  tomatoSlicedIsReceptacle::Hom(
    TomatoSliced, Receptacle)
  tomatoSlicedHasTemperature::Attr(
    TomatoSliced, Temperature)
  tomatoSlicedHasMass::Attr(TomatoSliced, Mass)
  tomatoSlicedHasMaterial::Attr(
    TomatoSliced, Material)

  Plate::Ob
  plateIsObject::Hom(Plate, Object)
  plateIsReceptacle::Hom(Plate, Receptacle)
  plateHasTemperature::Attr(Plate, Temperature)
  plateHasMass::Attr(Plate, Mass)
  plateHasMaterial::Attr(
    Plate, Material)

  Knife::Ob
  knifeIsObject::Hom(Knife, Object)
  knifeHasTemperature::Attr(Knife, Temperature)
  knifeHasMass::Attr(Knife, Mass)
  knifeHasMaterial::Attr(
    Knife, Material)
end
\end{minted}

\end{document}